\documentclass[conference]{ieeeconf}
\IEEEoverridecommandlockouts
\usepackage{cite}
\usepackage{amsmath,amssymb,amsfonts}
\usepackage{algorithmic}
\usepackage{graphicx}
\usepackage{textcomp}
\usepackage{xcolor}
\usepackage[dvipsnames]{xcolor}
\usepackage{soul}
\usepackage{siunitx}
\usepackage{caption}
\usepackage{subcaption}
\usepackage{booktabs}
\usepackage{float}
\usepackage{url}
\def\BibTeX{{\rm B\kern-.05em{\sc i\kern-.025em b}\kern-.08em
    T\kern-.1667em\lower.7ex\hbox{E}\kern-.125emX}}

\begin{document}

\title{\LARGE \bf {Whole-Body Model Predictive Control \\
for Spin-Aware Quadrupedal Table Tennis\\}
\thanks{}
}

\author{David Nguyen$^{1, 3}$, Zulfiqar Zaidi$^{2, 3}$, Kevin Karol$^{3}$, Jessica Hodgins$^{3}$, Zhaoming Xie$^{3}$
\thanks{$^{1}$Department of Mechanical Engineering, Massachusetts Institute of Technology, Cambridge, MA 02139, USA}
\thanks{$^{2}$Department of Mechanical Engineering, Georgia Institute of Technology, Atlanta, GA 30332, USA}
\thanks{$^{3}$RAI Institute, Cambridge, MA 02142, USA}
}

\maketitle


\begin{abstract}

Developing table tennis robots that mirror human speed, accuracy, and ability to predict and respond to the full range of ball spins remains a significant challenge for legged robots. To demonstrate these capabilities we present a system to play dynamic table tennis for quadrupedal robots that integrates high speed perception, trajectory prediction, and agile control. Our system uses external cameras for high-speed ball localization, physical models with learned residuals to infer spin and predict trajectories, and a novel model predictive control (MPC) formulation for agile full-body control. Notably, a continuous set of stroke strategies emerge automatically from different ball return objectives using this control paradigm. We demonstrate our system in the real world on a Spot quadruped, evaluate accuracy of each system component, and exhibit coordination through the system's ability to aim and return balls with varying spin types. As a further demonstration, the system is able to rally with human players.
\end{abstract}


\section{Introduction}

Table tennis is a fast-paced sport, requiring split second perception, prediction, strategizing, and response. 
In a competitive table tennis game, a player can move up to \SI{2.25}{\meter} in less than a second to accurately hit a \SI{4}{\centi\meter} diameter ball with a \SI{15}{\centi\meter} diameter paddle, all while making strategic decisions in fractions of a second~\cite{pingpong_speed}. 


For a robot to accurately control the trajectory and spin of a ping pong ball, it must solve four problems: ball perception, trajectory prediction, aiming, and swinging. First, the robot must accurately localize the ball which can travel at up to \SI{10}{\meter\per\second} with \SI{600}{\radian\per\second} of spin~\cite{pingpong_ball_speed, pingpong_ball_spin_speed}. Next, to know where and when to strike the ball, the robot needs to anticipate its motion and estimate spin. To hit a ball on this predicted trajectory to a desired landing location with a specific spin, a planner must create different swing types to strike the ball with the appropriate speed and angle. Finally, to execute this plan the robot must control its joints to ensure an accurate strike. Solving this series of problems at the speed of table tennis makes precise ball control an excellent case study for dynamic robotic control.

Existing table tennis robot systems generally consist of a robotic arm either attached to a fixed base, e.g.,~\cite{mit_pingpong, pingpong_rl_icra2021, pingpong_rl_tro2022}, limiting their range of movement, or include a customized fast-moving gantry, e.g.,~\cite{google_pingpong_system_rss}, which reduces the challenge of agile motion control at the expense of generalizability across robot platforms. In contrast, quadrupedal or bipedal legged robots must move with constraints similar to those of a human, trading off high speed motion and active balance. 


\begin{figure}
    \centering
    \includegraphics[width=0.95\linewidth, trim={0cm 5cm 0cm 7cm},clip]{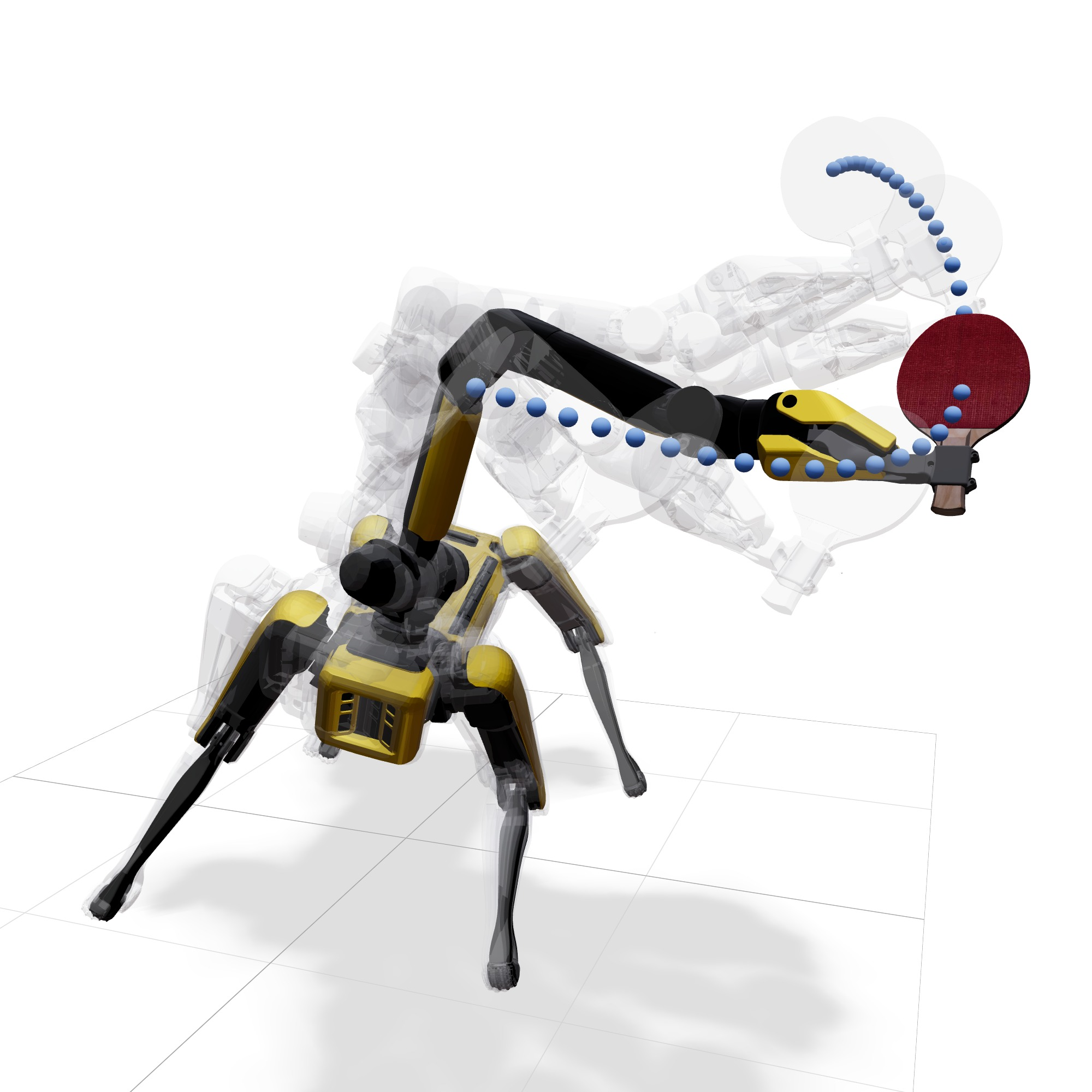}
    \caption{Dynamic whole-body quadruped swinging. Robot states are shown in gray with the paddle trajectory in blue. All renderings are generated using Viser \cite{viser}}
    \label{fig:money_shot}
\end{figure}

Our system uses a Boston Dynamics Spot equipped with a six degree of freedom (DoF) arm (Fig.~\ref{fig:money_shot}) and addresses the core challenges for robotic table tennis. We evaluate the accuracy of our ball localization and prediction systems using ground truth position data from a Vicon motion capture system, and ground truth spin data from a Spinsight Elite~\cite{pingpong_ball_spin_speed}. We assess our model predictive controller performance on hardware with demonstrations of returning balls across a range of speeds and spins while aiming at three different targets locations. To achieve these spins and positions, the controller exhibits emergent behavior that mirrors stroke strategies common in human players. Finally, we validate our controller on hardware by rallying with our system.

Our key contribution is the introduction of a Spot quadruped table tennis system capable of handling and generating competitive spin. Within this system, we developed a novel MPC formulation that can handle the continuous constraints of swing planning. We present our full system design, which consists of the following:

\begin{itemize}
    \item High-speed perception to accurately localize the ball

    \item Trajectory prediction to estimate ball state and trajectory

    \item Aiming planner to choose a paddle state from a desired target and outgoing spin.

    \item Quadruped MPC controller to strike the incoming ball
\end{itemize}


Notably, our system is able to return incoming spin of up to \SI{280}{\radian\per\second} and impart outgoing spin of up to \SI{200}{\radian\per\second}, surpassing prior work, e.g.,~\cite{liu2012racket, wang2023table}.
\section{Related Work}

\subsection{Robot Racket Sports}

Robots playing sports is a common challenge to foster robotics research to match the dynamic motion of humans and animals. Racket sports especially have become a popular test bed because of the need for agility and planning. Prior work has developed robot systems to play tennis~\cite{gatech_tennis,varsm_tennis,krishnatennis_2022, büchler2020} and badminton~\cite{eth_badminton, pingpong_rl_icra2025}. Because of the pace of the game, table tennis stands out within racket sports as a preferred robotics research challenge.

\subsection{Perception in Table Tennis Robot}
Prior works in robotic table tennis perception primarily address ball localization \cite{tebbe2019table,pingpong_detection_2019,Qingyu2025_icra,google_pingpong_system_rss,pingpong_detection_2012}, trajectory prediction \cite{tebbe2019table,g1_pingpong,Qingyu2025_icra,tebbe2020spin}, and spin estimation \cite{Qingyu2025_icra,tennis_detection_2024,tebbe2020spin}. Like some existing systems \cite{google_pingpong_system_rss,pingpong_detection_2019,tebbe2020spin}, we employ a pair of high-speed RGB cameras for ball localization. Many simpler trajectory prediction methods, however, disregard ball spin \cite{g1_pingpong,gatech_tennis}. This omission is problematic because ball flight, and contact dynamics are significantly influenced by spin \cite{contact_model} making it a critical component of the game. Some prior works estimate spin based on ball trajectory \cite{tebbe2020spin,tebbe2019table}. Alternatively, Gossard et al.~\cite{gossard2023spindoe} track non-regulation markings on the ball for precise spin estimation, which is incompatible with standard play. Other methods infer spin from human player poses \cite{tennis_detection_2024}, but this approach relies on noisy human pose measurements. Our system builds upon the methods of Tebbe et al.~\cite{tebbe2020spin} by adding a learned neural network to the model-based estimate from trajectory curvature. This approach improves future state prediction accuracy and handles incoming spin without the need for ball markings or human position estimation.

\subsection{Control in Table Tennis Robot}
To play table tennis effectively, the robot needs to perform accurate swing motions with high agility. Most prior works focus on control systems for a fully actuated robot arm that is either fixed or attached to a fast moving gantry. Systems often use human demonstration to construct motion primitives~\cite{pingpong_IJRR_2013} or as training data for imitation learning~\cite{pingpong_diffusion_icra2025}. 
Reinforcement learning algorithms, e.g.,~\cite{google_pingpong, pingpong_rl_tro2022, pingpong_rl_icra2025} are often employed to synthesize table tennis skills, but require large quantities of simulated or real world data to perform effectively. These techniques also require separate controllers for categorized shots~\cite{google_pingpong} because of the motion diversity of table tennis swings. Model-based methods also demonstrate effective control synthesis, e.g.,~\cite{mit_pingpong, ping_pong_RAS_2018}, without the need for data and can achieve greater shot diversity with a single controller. Our controller extends model-based methods for robot table tennis to handle the challenges presented by legged robots including balance.



Concurrent to our work, \cite{g1_pingpong} uses reinforcement learning to train a humanoid robot to play table tennis. In contrast to their reliance on using human motion capture data to bootstrap the stroke strategies, we demonstrate a wide range of stroke strategies common in human players that automatically emerges from solving MPC. We also demonstrate that our control system can both handle incoming spin and generate it using swings such as loops and chops which add top and back spin to the ball respectively.
\section{System}



\begin{figure}
    \centering
    \includegraphics[width=\linewidth]{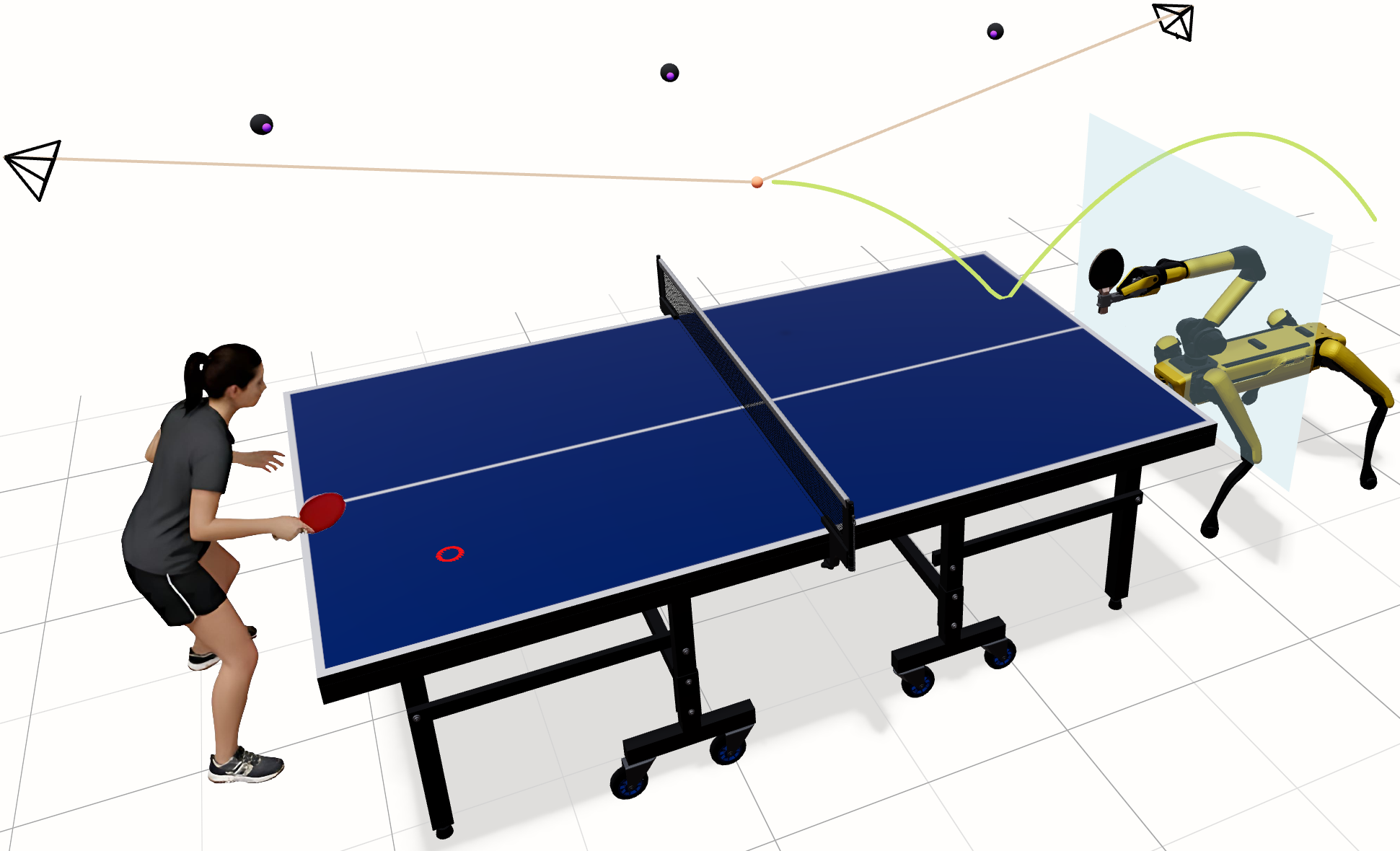}
    \caption{System diagram with RGB cameras shown as wire frame pyramids that detect the ball in orange. Its predicted trajectory is shown in green with the strike plane in blue. 
    The black motion capture cameras, located in the background, observe the position of the robot. The target ball landing location is in red on the table.
    }
    
    \label{fig:system_rendering}
\end{figure}

In this section, we describe our system (shown in Figure~\ref{fig:system_rendering}), including the perception, prediction, aiming, and control subsystems.

\subsection{Ball Detection and Localization}
The perception system is responsible for detecting and localizing the table tennis ball in 3D space. This task is accomplished through a stereo camera setup, a camera calibration procedure, and a high-speed detection pipeline.

\subsubsection{Camera Setup and Calibration}
The system utilizes two Power over Ethernet (PoE) RGB cameras (Lucid Arena), each with a resolution of 1400x1080 pixels, capturing images at 165 frames per second (fps). The cameras are mounted on the ceiling at opposite ends of the table and angled downwards to view the entire playing surface (see Figure \ref{fig:system_rendering}). To enable 3D reconstruction, both intrinsic and extrinsic camera parameters are needed. The intrinsic parameters are obtained from the specifications of the manufacturer. The extrinsic parameters are computed relative to a world coordinate frame whose origin is fixed at the center of the table. This calibration is achieved through a semi-automated process where key points corresponding to the table's corners are manually identified in a static image from each camera. Given the known 3D coordinates of these corners in the world frame and their corresponding 2D pixel coordinates, the extrinsic calibration (rotation and translation) for each camera is calculated by solving the Perspective-n-Point (PnP) problem~\cite{pnp}.

\begin{figure}
    \centering
    \includegraphics[width=\linewidth]{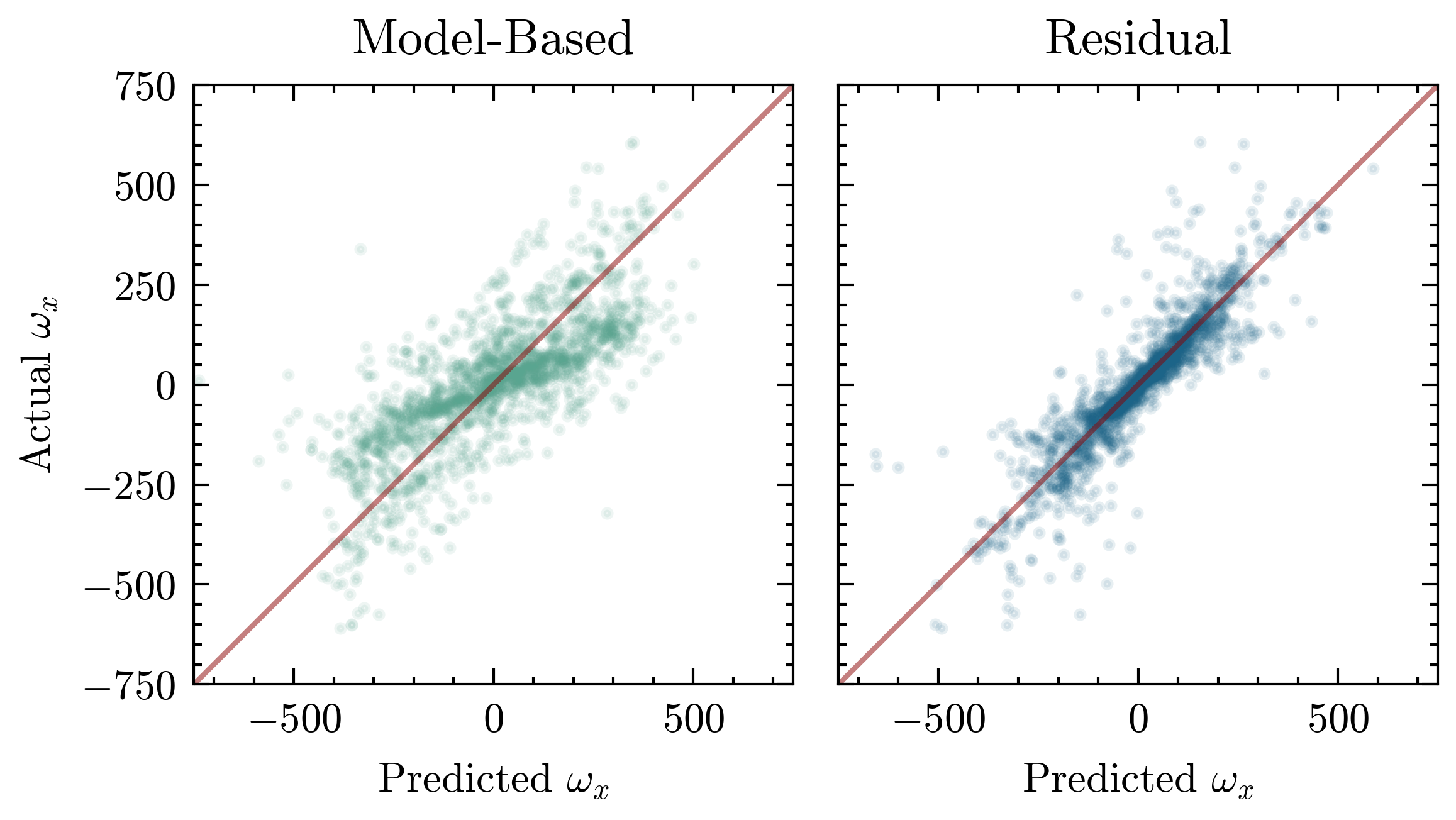}
    \caption{Spin predicton performance using the model-based estimate on the left and the residual network on the right.}
    \label{fig:spin_estimation_results}
\end{figure}

\subsubsection{Ball Detection and Localization}
Our ball detection pipeline is a two-stage process designed for high-speed performance, similar to the approach in \cite{gatech_tennis}.

\begin{enumerate}
    \item \textbf{Motion Detection:} For each incoming image stream, a background subtraction algorithm first identifies dynamic regions of the image. This operation isolates moving objects, primarily the ball, from the static background.
    \item \textbf{Object Detection:} The segmented moving regions are then composited into a smaller 480x480 pixel image patch. A fine-tuned YOLO convolutional neural network (CNN)~\cite{yolov3,darknet_yolo} is applied to this composite patch to detect the ball's 2D pixel coordinates. This approach significantly reduces computational load by avoiding the need to run the CNN on the full-resolution image. The YOLO network was refined on a custom dataset of these composite patches to handle the domain shift from standard datasets.
\end{enumerate}

Once the ball is detected in both camera streams at pixel coordinates $(u_1, v_1)$ and $(u_2, v_2)$, its 3D position $P = [X, Y, Z]^T$ in the world coordinate frame is calculated using stereo triangulation and passed to the prediction system.

\subsubsection{Performance}
With cameras operating at 165\,fps (an interval of ~6\,ms per frame), the entire detection process is completed in approximately 4\,ms (2.5\,ms for background subtraction and 1.5\,ms for CNN inference). This low latency ensures that a detection result is available well before the next frame is captured.

\subsection{Ball Trajectory Prediction and Spin Estimation}

\newcommand\vel{\mathbf{v}}
\newcommand\cdrag{C_D}
\newcommand\cmagnus{C_M}

To predict when and where the robot must strike the ball, we require an estimate of the ball's linear and angular velocity from ball position measurements. We used methods proposed by Tebbe et al. \cite{tebbe2020spin} for both ball velocity ($\vel$) and spin ($\omega$) estimation. This process includes fitting polynomials to the history of ball positions and taking their derivatives for an approximation of $\vel$. To estimate $\omega$, we sample a grid of these velocity points and construct a discrete approximation of the dynamics in \eqref{eq:finite_difference_ball_dynamics}.

\begin{equation}
    \frac{\vel_{i+1} - \vel_i}{\Delta t_i} \approx -\cdrag \|\vel_i\| \vel_i + \cmagnus(\boldsymbol{\omega} \times \vel_i) - \mathbf{g}
    \label{eq:finite_difference_ball_dynamics}
\end{equation}

where $\cdrag$ and $\cmagnus$ represent the lumped coefficients of the drag and Magnus effects respectively. By rearranging using the skew-symmetric matrix into equation \eqref{eq:skew_sym_simplification} and stacking for all $\vel_i$ from the original velocity grid, we can perform a least squares solve for $\omega$ assuming it stays unchanged throughout ball flight.

\begin{equation}
    \cmagnus \begin{bmatrix} \vel_i \end{bmatrix}_\times \boldsymbol{\omega} = \frac{\vel_{i+1} - \vel_i}{\Delta t_i} + \cdrag \|\vel_i\| \vel_i + \mathbf{g}\label{eq:skew_sym_simplification}
\end{equation}

This technique from Tebbe et al. \cite{tebbe2020spin} provides an estimate of spin given an analytical model, the performance of which is shown in Figure \ref{fig:spin_estimation_results} on the left. All ground truth spin values were captured using a Spinsight Elite ball tracking system alongside custom marked balls. This system was also used for the characterization of parameters $\cmagnus$ and $\cdrag$. 

To refine any unmodeled effects from our initial state estimate, we trained a residual network on 650 unique ball trajectories with varying spin. For data augmentation, we performed five random rotations about the z-axis and added them to the dataset. We then trained a small network with inputs $\omega$ from the least squares estimate and the polynomial coefficients to indicate curvature. Using $\text{R}^2$ as a performance metric, our learned residual improved the purely model-based approach from a 0.42 to 0.70 as seen in in Figure~\ref{fig:spin_estimation_results}. 

Using our state estimate, we integrate the ball trajectory given the same dynamics from \eqref{eq:finite_difference_ball_dynamics} along with the table rebound model from Nonomura et al. \cite{contact_model}. The integration stops when the ball reaches a fixed strike plane located \SI{0.5}{\meter} in front of the quadruped base. The terminal ball state is then reported to the aiming controller to choose a paddle state.

\subsection{Strike Aiming}

Using the anticipated ball state provided from the prediction system, we must find a contact paddle state that returns the ball to the other side of the table. This paddle state is described using $\mathbf{p}_\text{des}$, $\mathbf{v}_\text{des}$, and $\mathbf{n}_\text{des}$, the paddle positon, velocity, and face normal vector respectively. 

To choose these parameters, we designed a high-level aiming controller that takes in the desired landing position $\mathbf{p}_\text{land}$, spin $\omega^+$, and landing time $t_\text{land}$ and produces the desired paddle state. This problem is formulated as a constrained optimization in \eqref{eq:aiming_cost}-\eqref{eq:landing_constraint}.

\newcommand\ball{\mathbf{r}_\text{b}}
\newcommand\dball{\mathbf{\dot{r}}_\text{b}}
\newcommand\pdes{\mathbf{p}_\text{des}}
\newcommand\ndes{\mathbf{n}_\text{des}}
\newcommand\vdes{\mathbf{v}_\text{des}}
\newcommand\ballstate{\mathbf{x}_\text{b}}

\begin{subequations}
    \begin{align}
        \min_{\ball, \dball, \ndes, \vdes} \quad & W_v\|\mathbf{v}_\text{des}\|_2^2 \label{eq:aiming_cost} \\
        \text{s.t.} \quad & \ball[0] = \pdes \label{eq:ball_pos} \\
        & \begin{bmatrix}
            \dball[0] \\
            \boldsymbol{\omega}^+
        \end{bmatrix} = f_\text{contact}(\dball^-, \vdes, \ndes, \boldsymbol{\omega}^-) \label{eq:contact_dyn} \\
        & \ballstate[n+1] = \ballstate[n] + f_\text{aero}(\ballstate[n])\Delta t \ \forall n \label{eq:aero_dyn} \\
        & \ball[N-1] = \mathbf{p}_\text{land} \label{eq:landing_constraint}
    \end{align}
\end{subequations}

\noindent
where $\ball$ and $\dball \in \mathbb{R}^{3 \times N}$ are the ball positions and velocities post collision and $N$ is the number of trajectory nodes. $f_\text{contact}$ represents the paddle and ball contact dynamics which are a modified version of those in \cite{contact_model} using vector operations rather than rotation matrices. This simplified the number of instructions in the formulation in contrast to using two sets of frame rotations. $f_\text{aero}$ corresponds to the discrete aerodynamics in \eqref{eq:finite_difference_ball_dynamics} and $\ballstate = [\ball^\intercal, \dball^\intercal, {\boldsymbol{\omega}^+}^\intercal]^\intercal$ signifies the full ball state for simplified notation.



To solve this optimization problem in real time, we implemented Sequential Quadratic Programming (SQP) where we take a quadratic approximation of our cost with linearized constraints and solve that problem iteratively to handle any nonlinearity of the original problem. For our local QP solver, we utilize OSQP~\cite{osqp} and perform four SQP iterations before using the solution. All optimization formulations in this paper were generated using CasADi for fast function evaluation \cite{casadi}. 

\begin{figure}
    \centering
    \includegraphics[width=\linewidth]{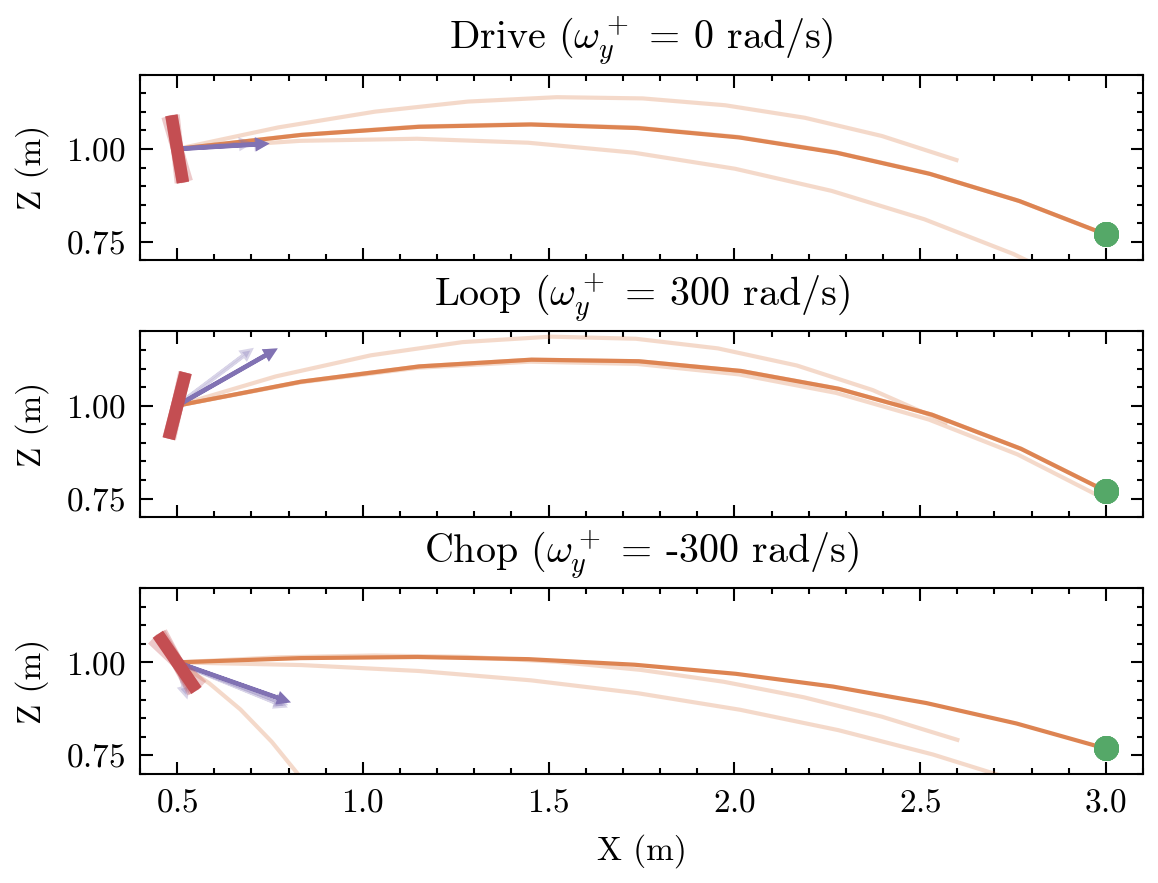}
    \caption{Convergence of aiming planner to a given $\mathbf{p}_\text{land}$, $\omega^+$, and $t_\text{land}$. The paddle orientation and velocity is shown in red and purple respectively. The simulated resulting trajectory is shown in orange and $\mathbf{p}_\text{land}$ is indicated with the green point. The lighter colored components represent intermediate solutions during the SQP iterations.}
    \label{fig:aiming_convergence}
\end{figure}

The convergence of the controller to an accurate numerical solution is shown in Figure~\ref{fig:aiming_convergence} where three shot types were tested: a flat paddle drive, a top spin loop, and back spin chop. For each of the displayed solutions, the inputs to the optimization problem were constant other than $\boldsymbol{\omega}^+$ which resulted in drastically different solutions for $\mathbf{v}_\text{des}$ and $\mathbf{n}_\text{des}$. The orange path of the ball is a simulated ball trajectory given the paddle state solutions during solve convergence. In all three cases, the controller converges to an accurate solution within four iterations which takes only \SI{1}{\milli\second}.

\begin{figure*}[t!]
     \centering
     \begin{subfigure}[b]{0.18\textwidth}
         \centering
         \includegraphics[width=\textwidth, trim={9cm 14cm 20cm 0cm},clip]{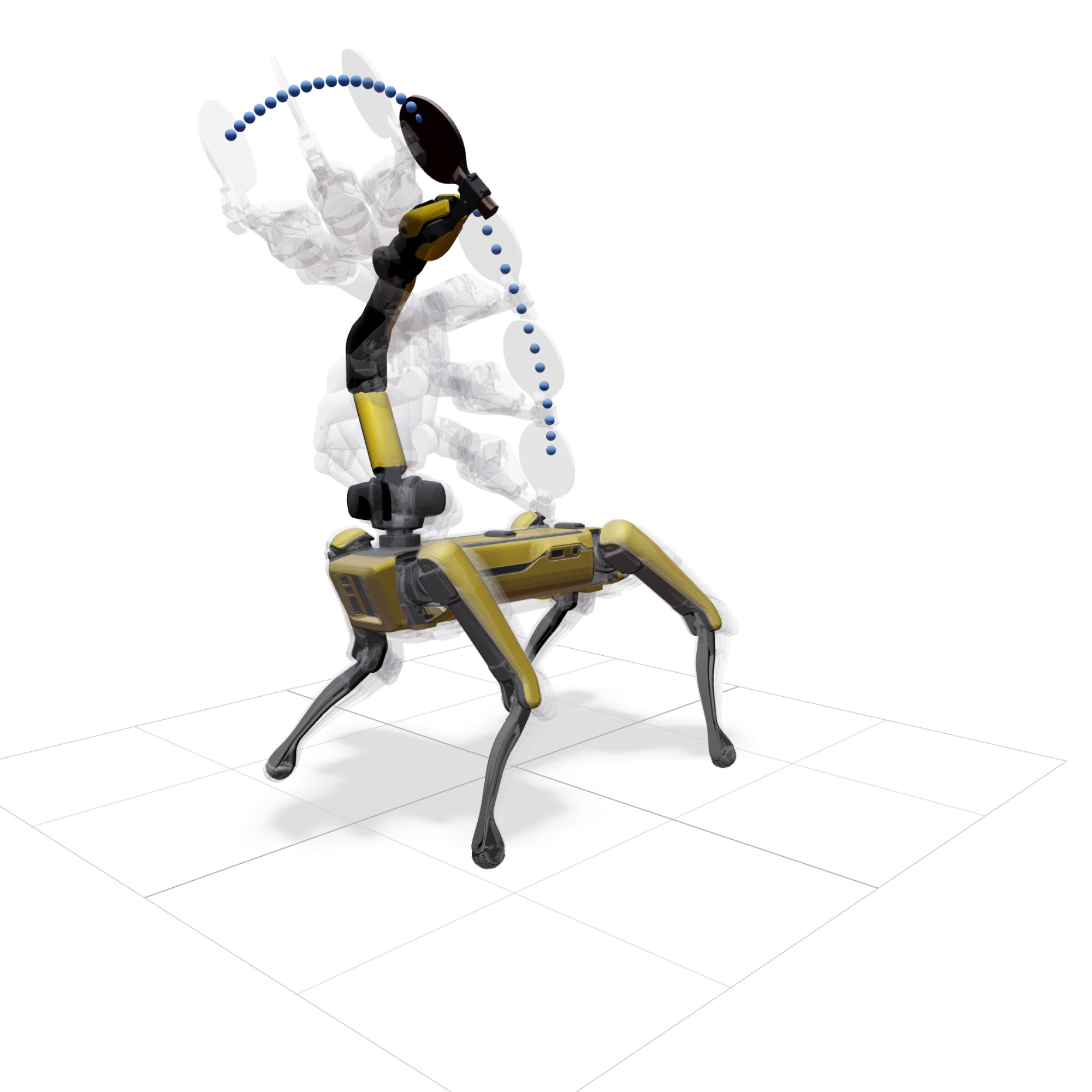}
         \caption{High Loop}
         \label{fig:top_spin_render}
     \end{subfigure}
     \hfill
     \begin{subfigure}[b]{0.18\textwidth}
         \centering
         \includegraphics[width=\textwidth, trim={9cm 14cm 20cm 0cm},clip]{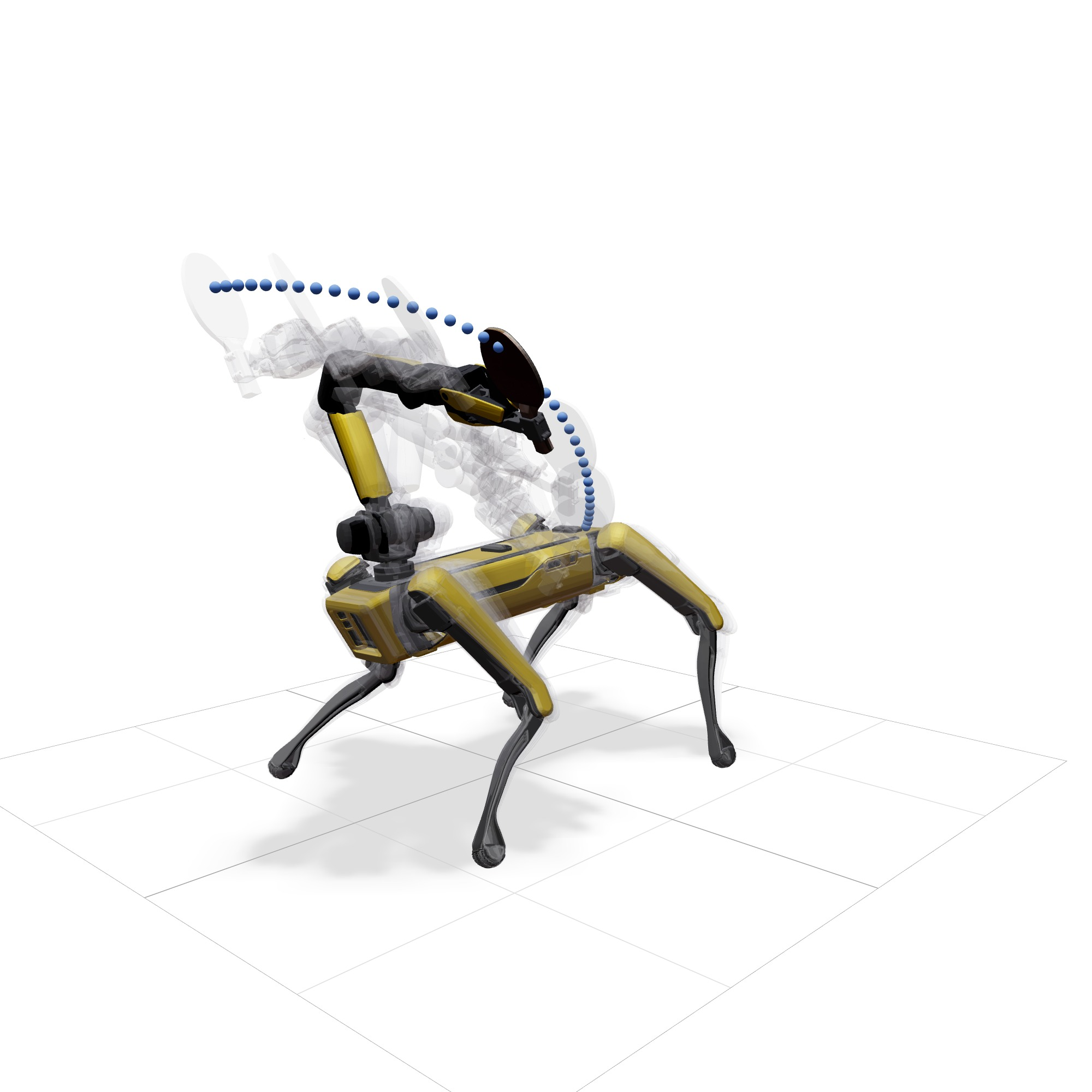}
         \caption{Loop}
         \label{fig:back_spin_render}
     \end{subfigure}
    \hfill
    \begin{subfigure}[b]{0.18\textwidth}
         \centering
         \includegraphics[width=\textwidth, trim={13cm 14cm 16cm 0cm},clip]{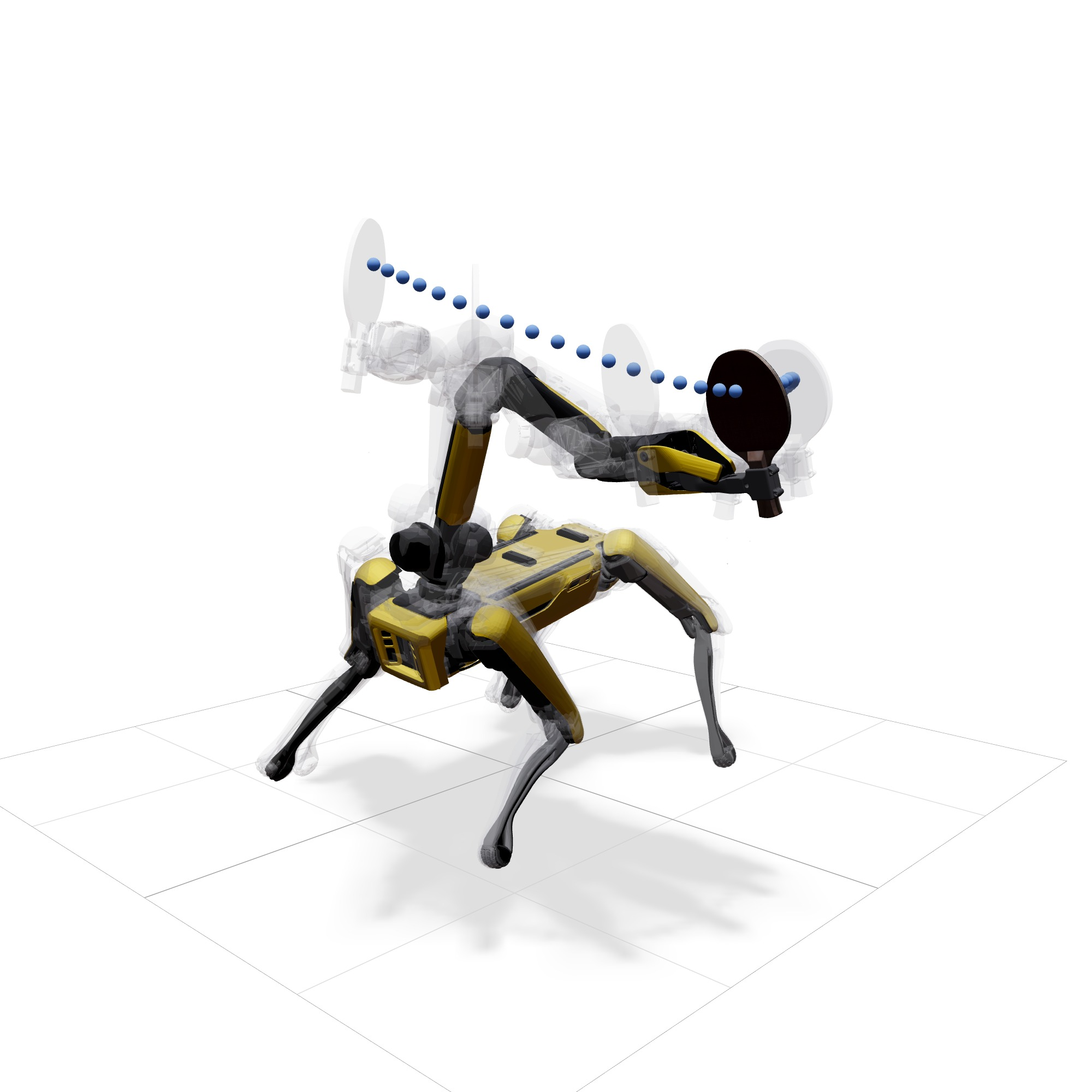}
         \caption{Far Drive}
         \label{fig:back_spin_render}
     \end{subfigure}
    \hfill
    \begin{subfigure}[b]{0.18\textwidth}
         \centering
         \includegraphics[width=\textwidth, trim={9cm 14cm 20cm 0cm},clip]{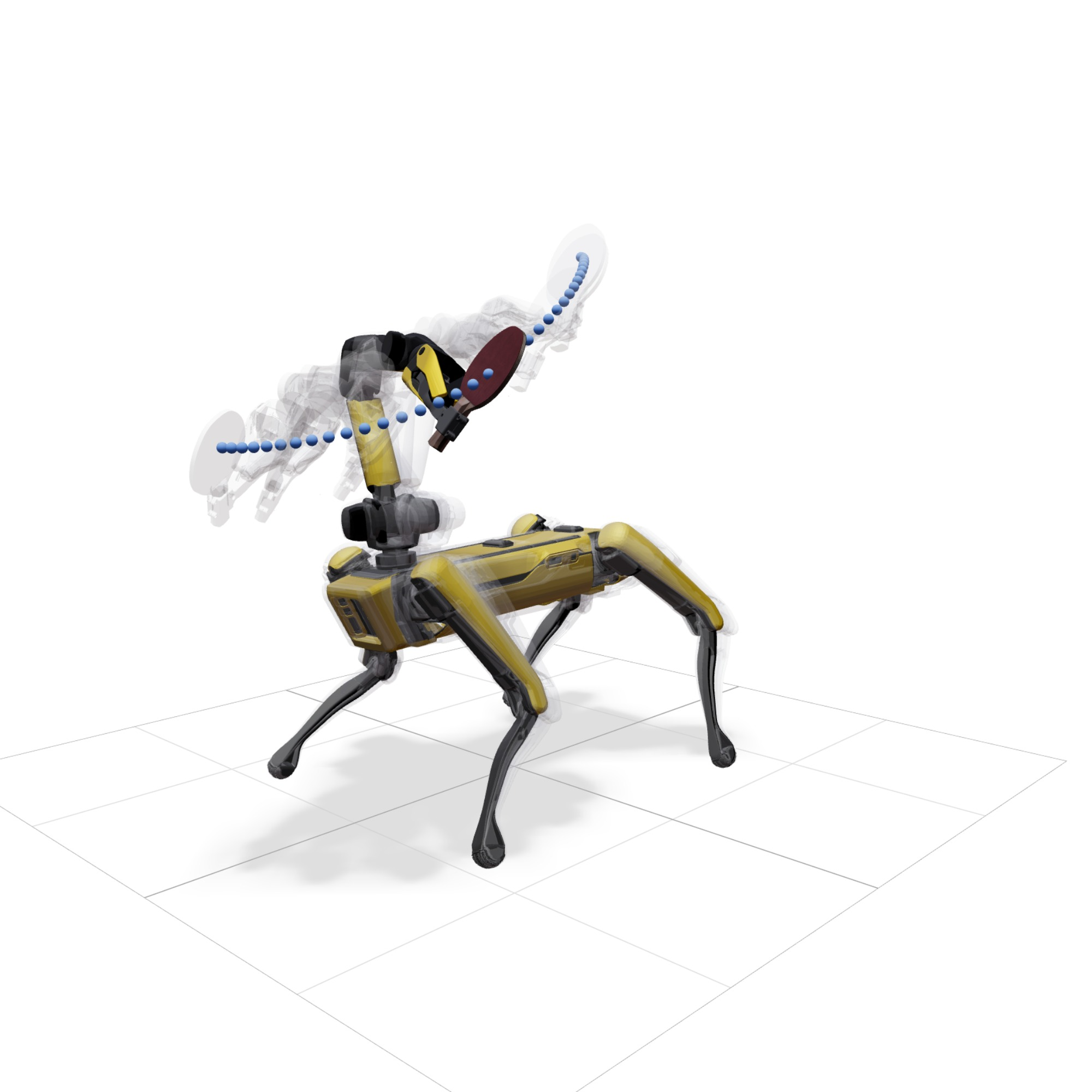}
         \caption{Chop}
         \label{fig:back_spin_render}
     \end{subfigure}
    \hfill
    \begin{subfigure}[b]{0.18\textwidth}
         \centering
         \includegraphics[width=\textwidth, trim={9cm 14cm 20cm 0cm},clip]{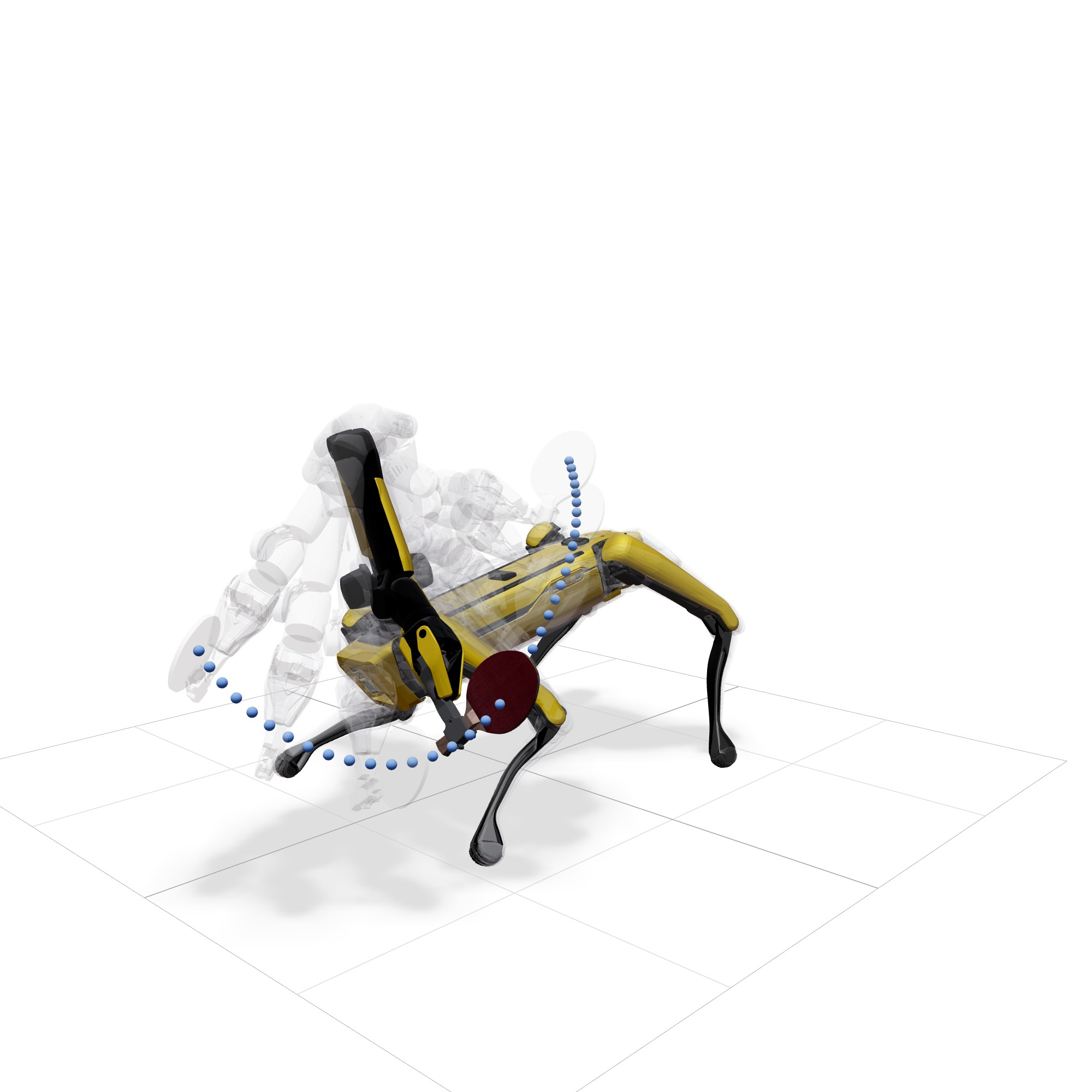}
         \caption{Low Chop}
         \label{fig:}
     \end{subfigure}
         \caption{Variety of swing types including loop (top spin), drive (no spin), and chop (back spin).}
        \label{fig:swing_variety}
\end{figure*}

\subsection{Whole Body Model Predictive Control}

\newcommand\p{\mathbf{q}_\text{c}}
\newcommand\pdd{\mathbf{\ddot{q}}_\text{c}}
\newcommand\qdd{\mathbf{\ddot{q}}}
\newcommand\qd{\mathbf{\dot{q}}}
\newcommand\q{\mathbf{q}}
\newcommand\qs{\mathbf{q}_\text{s}}
\newcommand\qds{\mathbf{\dot{q}}_\text{s}}
\newcommand\B{\mathcal{B}}
\newcommand\fk{\mathcal{K}}

To achieve the desired paddle state from the aiming controller and prediction system, we designed a model-based kinematic planner coupled with a whole-body controller that generates dynamically feasible swings.

\subsubsection{Kinematic Planner} To generate a swinging motion that can adapt to changing strike conditions and plan for a return trajectory, we require a constraint that starts at the end of our planning horizon and moves closer to the beginning as we execute our motion. This proves challenging for MPC formulations like multiple shooting or direct collocation because of their discrete dynamics and the continuous nature of this strike constraint. For this reason, our kinematic planner uses parametric Bezier curves rather than discrete nodes. This allows for a single strike constraint that can be enforced anywhere along the planning horizon.

Equations \eqref{eq:kin_mpc_cost}-\eqref{eq:kin_ori} showcase this optimization formulation with $\p \in \mathbb{R}^{24 \times 8}$ representing the Bezier curve control points for each joint. In the equation, the Bezier curve result is calculated using function $\B$ with the time and Bezier curve parameters as inputs. 

\begin{subequations}
    \begin{align}
        \min_{\p, \pdd, \q_\text{s}, \qd_\text{s}} \quad & W_a f_a(\pdd) + W_r f_r(\q_\text{rest}, \p) \label{eq:kin_mpc_cost} \\
        \text{s.t.} \quad & \B(\p, 0) = \q_0 \label{eq:kin_init_pos} \\
        & \B'(\p, 0) = \qd_0 \label{eq:kin_init_vel} \\
        & \q_\text{min} \le \B(\p, t) \le \q_\text{max} \ \forall t \in (0, t_f] \label{eq:kin_jnt_lim} \\
        & \fk_f(\B(\p, t)) = \fk_f(\q_0) \ \forall t \in (0, t_f] \label{eq:kin_foot} \\
        & \fk_p(\qs) = \mathbf{p}_\text{des} \label{eq:kin_pos} \\
        & \mathbf{J}_p(\qs)\qds = \mathbf{v}_\text{des} \label{eq:kin_vel} \\
        & \|\fk_n(\qs) - \fk_p(\qs) - \mathbf{n}_\text{des} \|_2^2 \le \epsilon_n \label{eq:kin_ori}
    \end{align}
\end{subequations}


\noindent
Equations \eqref{eq:kin_init_pos} and \eqref{eq:kin_init_vel} ensure the planned trajectory starts at the current state of the robot while the joint limits are constrained in \eqref{eq:kin_jnt_lim}. $\fk_f$ represents the forward kinematics of each foot in equation \eqref{eq:kin_foot} which keeps them stationary throughout the swing. Both \eqref{eq:kin_jnt_lim} and \eqref{eq:kin_foot} are enforced at sampled points along the Bezier curves. Finally \eqref{eq:kin_pos}-\eqref{eq:kin_ori} drive the end effector to match the desired paddle strike conditions at the strike time $t_s$. Here, $\fk_p$ and $\fk_n$ are the forward kinematics for the center of the paddle and a point above the paddle face respectively.

To keep the end effector and other kinematic constraints decoupled, we add slack decision variables $\q_\text{s}$ and $\qd_\text{s} \in \mathbb{R}^{24}$ which are the joint and body state of the robot at strike. Equations \eqref{eq:kin_slack_pos} and \eqref{eq:kin_slack_vel} constrain these variables to lie along the planned Bezier curves at $t_s$.

\begin{subequations}
    \begin{align}
        \B(\p, t_s) = \qs \label{eq:kin_slack_pos} \\
        \B'(\p, t_s) = \qds \label{eq:kin_slack_vel}
    \end{align}
\end{subequations}

To shape the swing, cost function $f_a$ limits the robot's acceleration while $f_r$ keeps its position close $\q_\text{rest}$, the rest stance. Because of the low distortion between the Bezier control points and the curve they parametrize, we can formulate these functions simply using the decision variables to keep the problem quadratic like in equation \eqref{eq:pos_cost}.

\begin{equation}
    f_r(\q_\text{rest}, \p) = \sum_{n=0}^{N_c} (\p[n] - \q_\text{rest})^\intercal \mathbf{W}_\text{j} (\p[n] - \q_\text{rest}) \label{eq:pos_cost}
\end{equation}

\noindent
where $\mathbf{W}_\text{j} \in \mathbb{R}^{24 \times 24}$ is a diagonal weighting matrix on each joint. Although indirect, this cost function keeps all control points close to the rest position $\q_\text{rest}$ which regularizes $\p$ without the need of computing any curve points. Similarly, we can minimize an acceleration proxy through $f_a$ in equation \eqref{eq:acc_cost} alongside slack variable constraint \eqref{eq:acc_slack}.

\begin{subequations}
    \begin{align}
        f_a(\pdd) = \sum_{n=0}^{N_c-2} \pdd[n]^\intercal \mathbf{W}_\text{j} \pdd \label{eq:acc_cost} \\
        \pdd[n] = \p[n+2] - 2\p[n+1] + \p[n] \label{eq:acc_slack}
    \end{align}
\end{subequations}


Similar to the strike constraints, slack variables $\pdd$ are introduced here to keep the cost quadratic and remove any coupling terms between control points.



Together, both $f_a$ and $f_r$ create a simple convex cost function that regularizes the swinging motion. An additional benefit of this formulation is that we can also utilize this optimization problem for returning to our rest position. If we disable constraints \eqref{eq:kin_pos}-\eqref{eq:kin_ori} then the solution to our optimization problem becomes a smooth trajectory back to the rest position from our current state. Therefore, this one problem can both plan for swings and return trajectories. 

We use the same SQP solver as the aiming system to solve this swing optimization problem. Because our cost is already quadratic, this solver only handles the constraint nonlinearity through multiple solves. During execution, we solve five SQP iterations instead of solving to convergence which allows the controller to run at \SI{100}{\Hz}.

\subsubsection{Whole Body Controller}

With this kinematic motion planner, we can now generate future trajectories that will meet our strike conditions, but we require a method of generating feedforward torques that abide by our dynamics. This can be accomplished using a whole-body controller which finds feedforward torques $\mathbf{u}$ given dynamics constraints with the goal of achieving a desired acceleration $\qdd_\text{des}$. This optimization is formulated in \eqref{eq:whole_body_cost}-\eqref{eq:grf_constraint}.

\begin{subequations}
    \begin{align}
        \min_{\qdd, \mathbf{u}, \boldsymbol{\lambda}} \quad & \| \qdd - \qdd_\text{des} \|^2 \label{eq:whole_body_cost} \\
        \text{s.t.} \quad & \mathbf{M}(\q)\qdd + \mathbf{C}(\q, \qd) = \boldsymbol{\tau}_\text{g} + \mathbf{B}\mathbf{u} + \mathbf{J}^\intercal(\q)\boldsymbol{\lambda} \label{eq:dyn_constraint} \\
        & \boldsymbol{\lambda} \in \mathcal{FC}(\mu) \label{eq:grf_constraint}
    \end{align}
\end{subequations}

\noindent
where $\mathbf{M}$, $\mathbf{C}$, and $\boldsymbol{\tau}_\text{g}$ are the mass matrix, Coriolis terms, and gravity terms respectively of the general robotic manipulator equations. The ground reaction forces are also solved for and denoted by $\boldsymbol{\lambda} \in \mathbb{R}^{12}$ while $\mathbf{J}$ is the Jacobian of the robot feet with respect to $\q$. To keep the constraints linear, the friction cone $\mathcal{FC}$ is approximated using a pyramidal approach. Together, this problem is a simple Quadratic Program and easily solved with off-the-shelf solvers, in this case OSQP \cite{osqp}.

With this combination of kinematic MPC planning and the dynamic whole-body controller, we are able to execute dynamic swings on hardware with a re-planning frequency of \SI{100}{\hertz}. Five swings from this controller are shown in Figure \ref{fig:swing_variety}, each with the strike state shown along with the swing motion.
\section{Evaluations}

\subsection{Ball Localization}

\begin{figure}[h!]
    \centering
    \includegraphics[width=\linewidth]{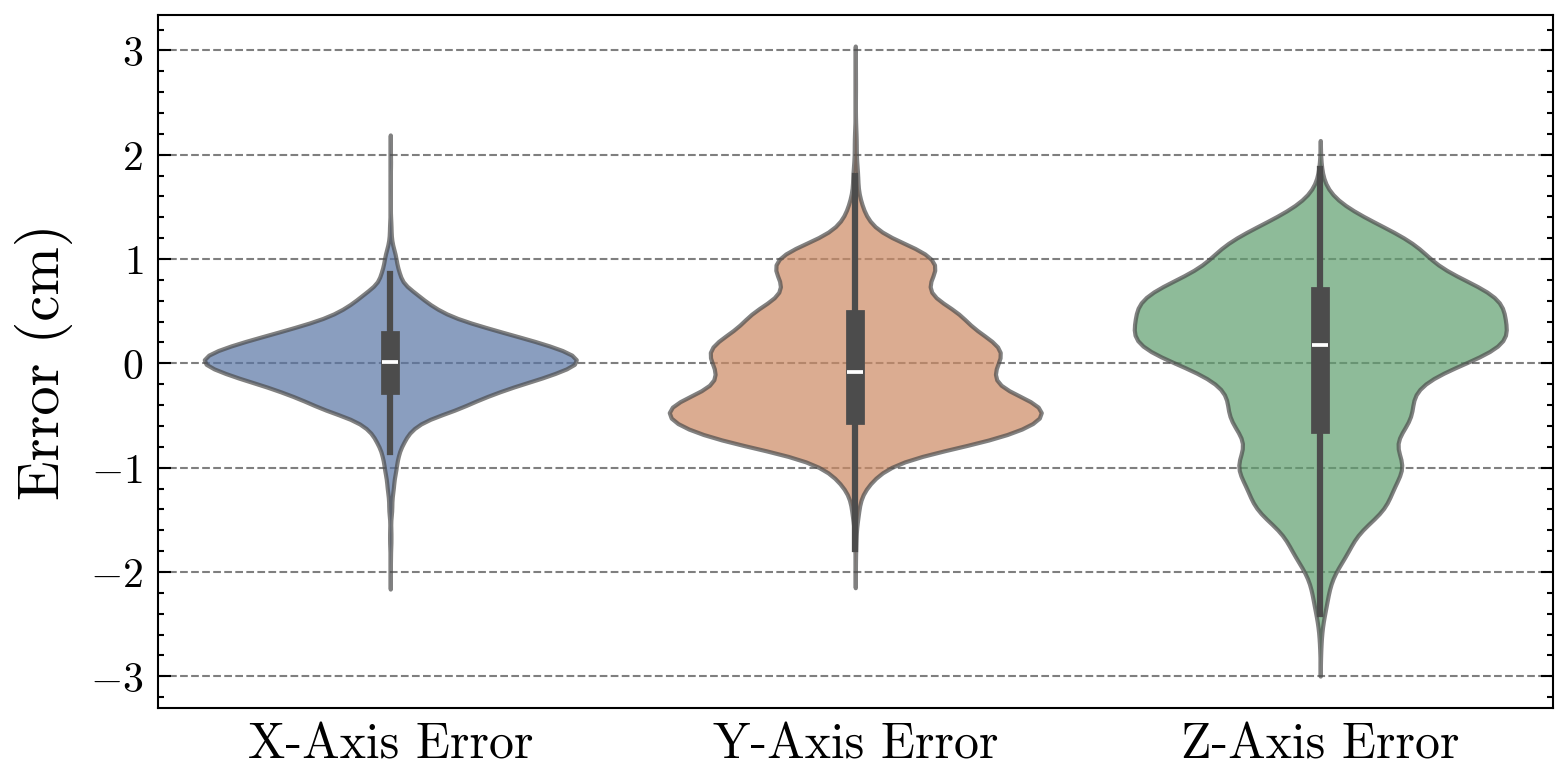}
    \caption{Distribution of the RGB position measurement error, calculated against data from a Vicon motion capture system.}
    \label{fig:ball_position_error}
\end{figure}

The performance of the RGB-based ball detection subsystem was quantitatively evaluated against ground truth Vicon motion capture data. A table tennis ball, outfitted with retro-reflective markers, was moved throughout the detection region while its position was tracked concurrently by both motion capture and our RGB detection systems. The resulting position measurement errors are presented as violin plots in Figure~\ref{fig:ball_position_error}. The distributions indicate that the median error for each axis remains below 1 cm, demonstrating the high accuracy of our perception module.

\subsection{Spin Estimation and Prediction}

\begin{figure}
    \centering
    \includegraphics[width=\linewidth]{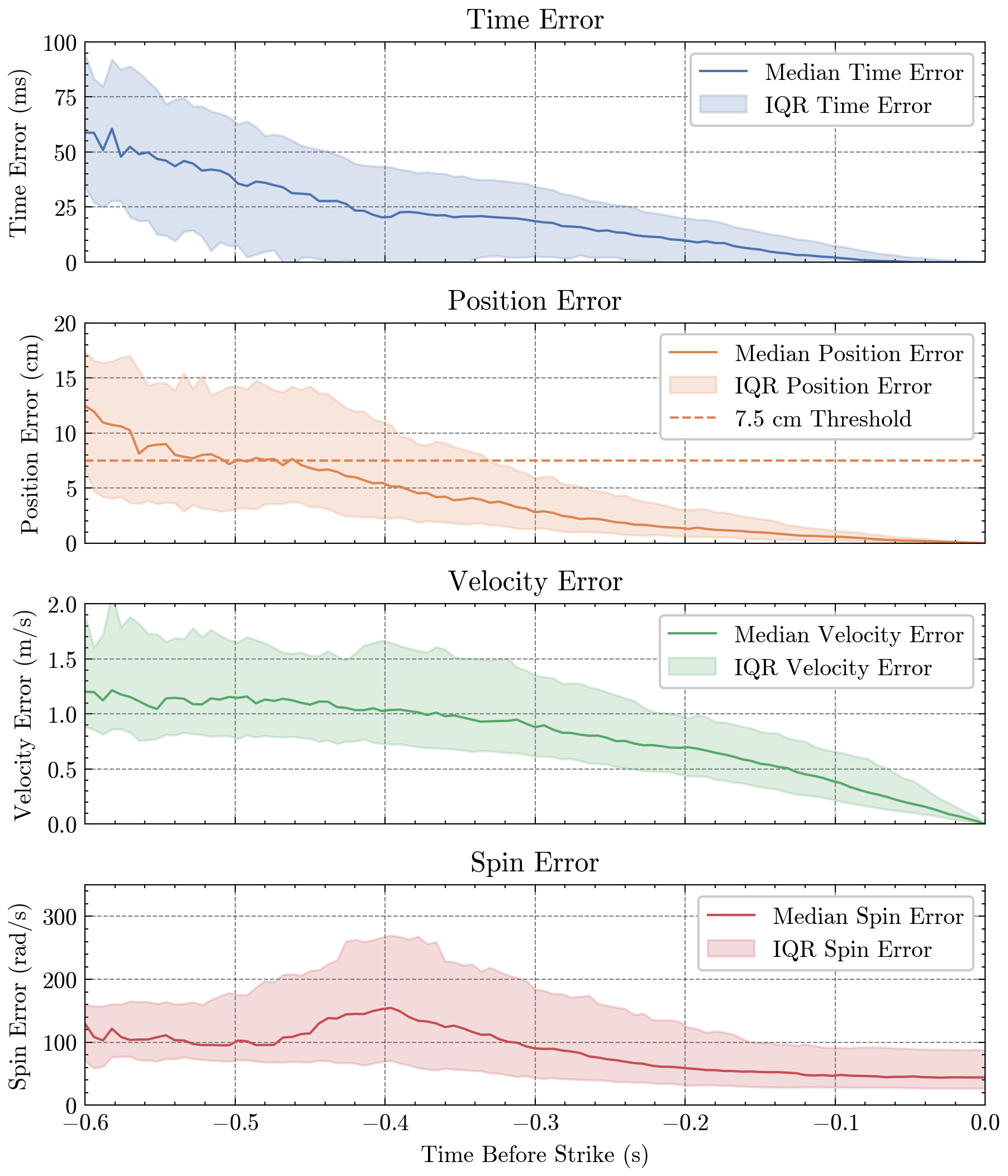}
    \caption{Prediction output errors through ball flight with median and inter-quartile range (IQR). The \SI{7.5}{\centi\meter} threshold for position error corresponds to the paddle radius, the minimum position accuracy require to strike the ball.}
    \label{fig:prediction_performance}
\end{figure}

Using a set of 600 recorded ball trajectories, we evaluated the performance of the prediction module by comparing the predicted and true final ball state as it approached the strike plane. Figure \ref{fig:prediction_performance} shows these errors in prediction timing, position, velocity, and spin.

The data indicates that our state estimate error reduces as more measured positions become available. Notably, the spin estimation converges to under \SI{55}{\radian\per\second} within \SI{150}{\milli\second} of the strike which provides time for the swing controller to adapt. The increase in spin estimation error at \SI{-0.4}{\second} can be explained by our system waiting for 30 ball detection points before estimating spin.




\subsection{Swing Controller}

\begin{figure}[t!]
    \centering
    \includegraphics[width=\linewidth]{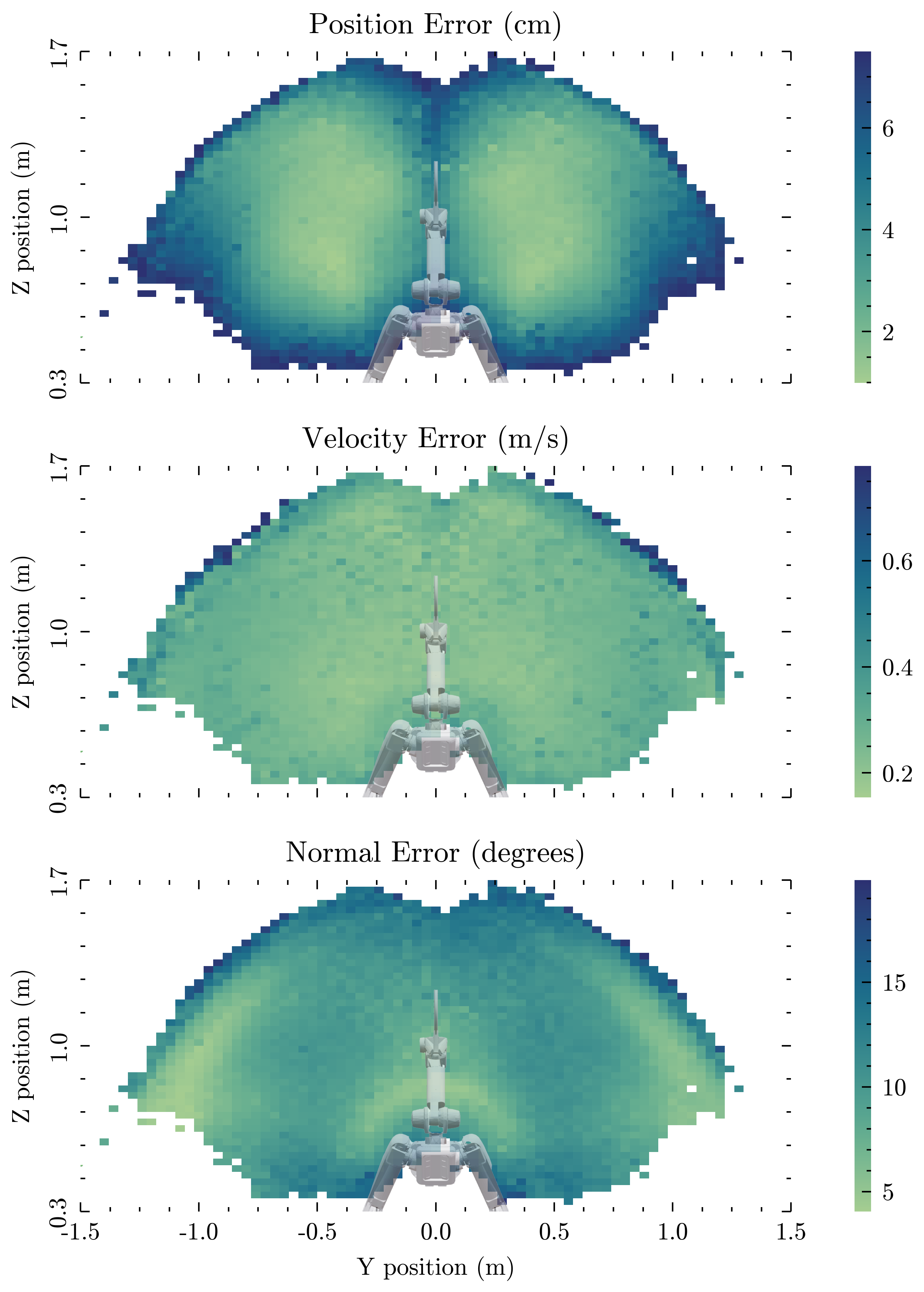}
    \caption{Position, velocity and orientation error of the paddle at strike tested with loop, chop, and drive shots at an array of positions.}
    \label{fig:mpc_sweep_heatmap}
\end{figure}

To understand how well our swinging controller can achieve arbitrary paddle states within the strike plane, we tested three types of swings at a dense grid of positions summing to 3750 samples. Each point was generated by simulating the kinematic planner and whole-body controller throughout a full swing while capturing the state of the robot at the time of strike. For each data point we then evaluated the paddle position, velocity, and orientation errors over the grid of tested positions as seen in Figure \ref{fig:mpc_sweep_heatmap}. A threshold of \SI{7.5}{\centi\meter}, \SI{1}{\meter\per\second}, and \SI{20}{\degree} for position, velocity, and orientation error was set to bound the heatmap regions which represent the strike workspace of our controller and robot. The position error heatmap indicates the best performance directly to the right and left of the robot with further and closer strikes being harder to reach. Within the position error bounds, the robot has velocity tracking error of under \SI{0.5}{\meter\per\second}. On the other hand,  the orientation error has a unique pattern with the best performance close and far from the robot. This result is likely related to our nonlinear and inequality orientation constraint which can be hard to solve for given the low number of SQP iterations during run time.

\subsection{System}

\begin{figure}
    \centering
    \includegraphics[width=\linewidth]{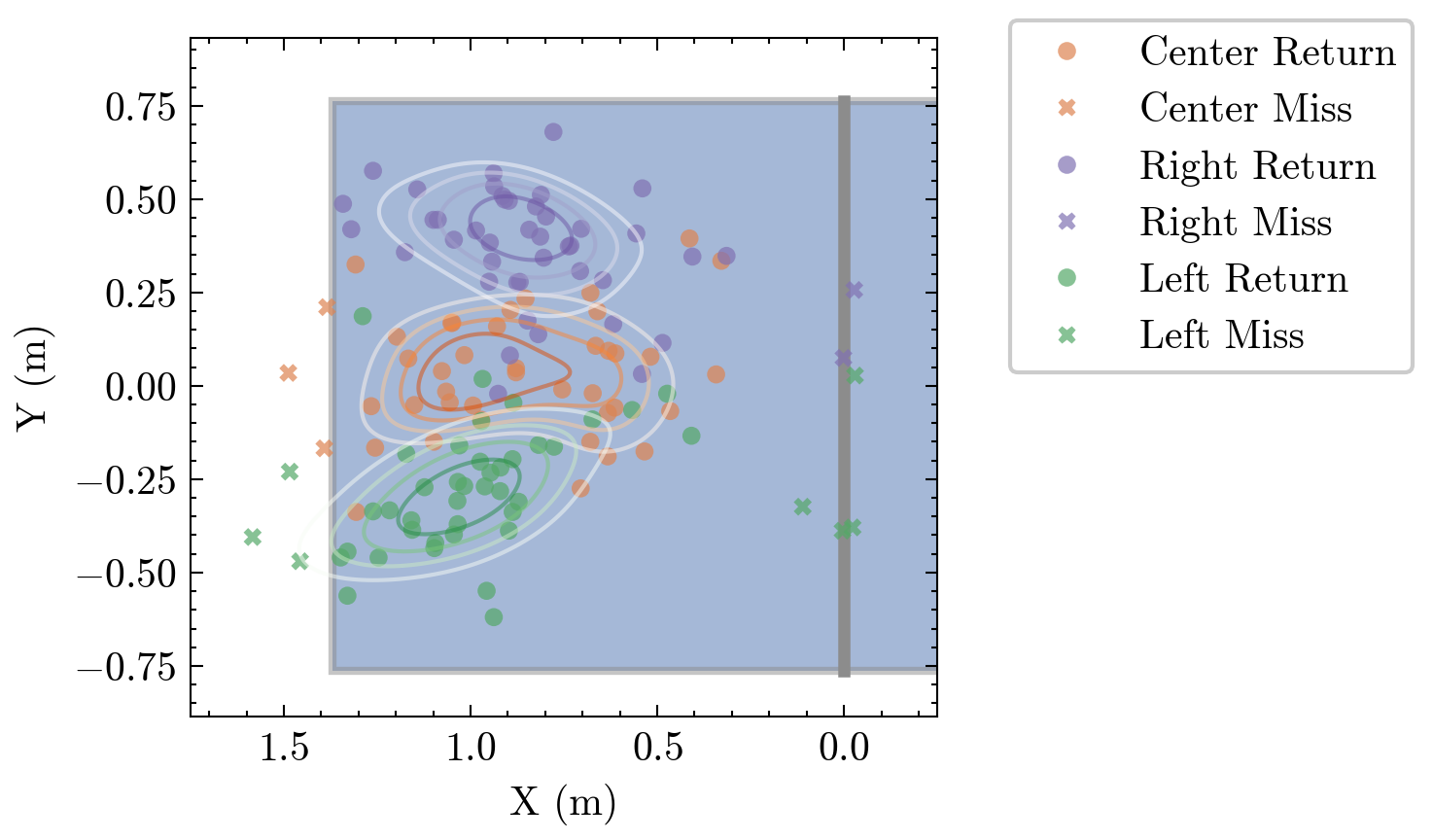}
    \caption{Landing locations when aiming to the right, center, and left. Shots originated from Spot on the right of the plot.}
    \label{fig:landing_locations}
\end{figure}

To evaluate how well the system works as a whole, we struck 150 balls and recorded their landing locations with three different $\mathbf{p}_\text{land}$ values. Figure~\ref{fig:landing_locations} includes the recorded landing locations for each target marked as a return or miss. Over the 150 trials, 90.1\% were returned with a clear aiming pattern given $\mathbf{p}_\text{land}$.

The state estimation and prediction system utilizes a residual network on top of a model-based estimator for $\boldsymbol{\omega}^-$. We tested the necessity for these components with a small ablation, which included no spin estimation, only model-based, and the full residual estimator. To choose the outgoing spin $\boldsymbol{\omega}^+$, we implemented a simple heuristic strategy used by players to return shots with the same spin as they are received. For each setup, we tested five different spin values shown in Table~\ref{tab:return_rate_diff_spin} and recorded the return rate over 25 trials.


\begin{table}[h]
    \centering
    \begin{tabular}{c c c c}
    & \multicolumn{3}{c}{\textbf{Spin Detection}} \\
    \cmidrule(lr){2-4}
    $\boldsymbol{\omega}_y^-$ & None & Least Squares & Residual\\
    \toprule
    220 rad/s & 0\% & 28\% & \bf{44\%}\\
    100 rad/s & 12\% & 52\% & \bf{72\%}\\
    0 rad/s & 72\% & 68\% & \bf{84\%}\\
    -125 rad/s & 12\% & 44\% & \bf{88\%}\\
    -280 rad/s & 40\% & 68\% & \bf{88\%}\\
    \midrule
    \bf{Mean} & 27.2\% & 52.0\% & \bf{75.2\%}\\
    \bottomrule
    \end{tabular}
    \caption{Return rate for different incoming ball spin $\boldsymbol{\omega}_y^-$ with changing degrees of spin detection. Negative $\boldsymbol{\omega}_y^-$ correspond to top spin while positive values indicate back spin.}
    \label{tab:return_rate_diff_spin}
\end{table}

This shows the need for the spin estimation and the improvement of return performance when the residual network is included. Overall, the full system was capable of handling a wide variety of ball spin with a mean return rate of 75\%.

Since the aiming system is capable of solving for the exiting ball spin $\boldsymbol{\omega}^+$, we tested the system's accuracy in generated this spin. Table~\ref{tab:spin_generation} includes the mean and standard deviation over 25 trials for four different target spin values.

\begin{table}[h]
    \centering
    \begin{tabular}{c c c}
    $\boldsymbol{\omega}_y^+$ & Mean & Std. Dev. \\
    \toprule
    200 rad/s & 191.9 rad/s & 16.5 rad/s \\
    125 rad/s & 118.1 rad/s & 15.0 rad/s \\
    -125 rad/s & -133.1 rad/s & 11.5 rad/s \\
    -200 rad/s & -182.6 rad/s & 24.0 rad/s \\
    \bottomrule
    \end{tabular}
    \caption{Mean and standard deviation of ball spin added by Spot for different desired spin speeds. Here, positive values correspond to top spin and negative to back spin.}
    \label{tab:spin_generation}
\end{table}

\begin{figure}[h!]
     \centering
     \begin{subfigure}[b]{0.45\textwidth}
         \centering
         \includegraphics[width=\textwidth]{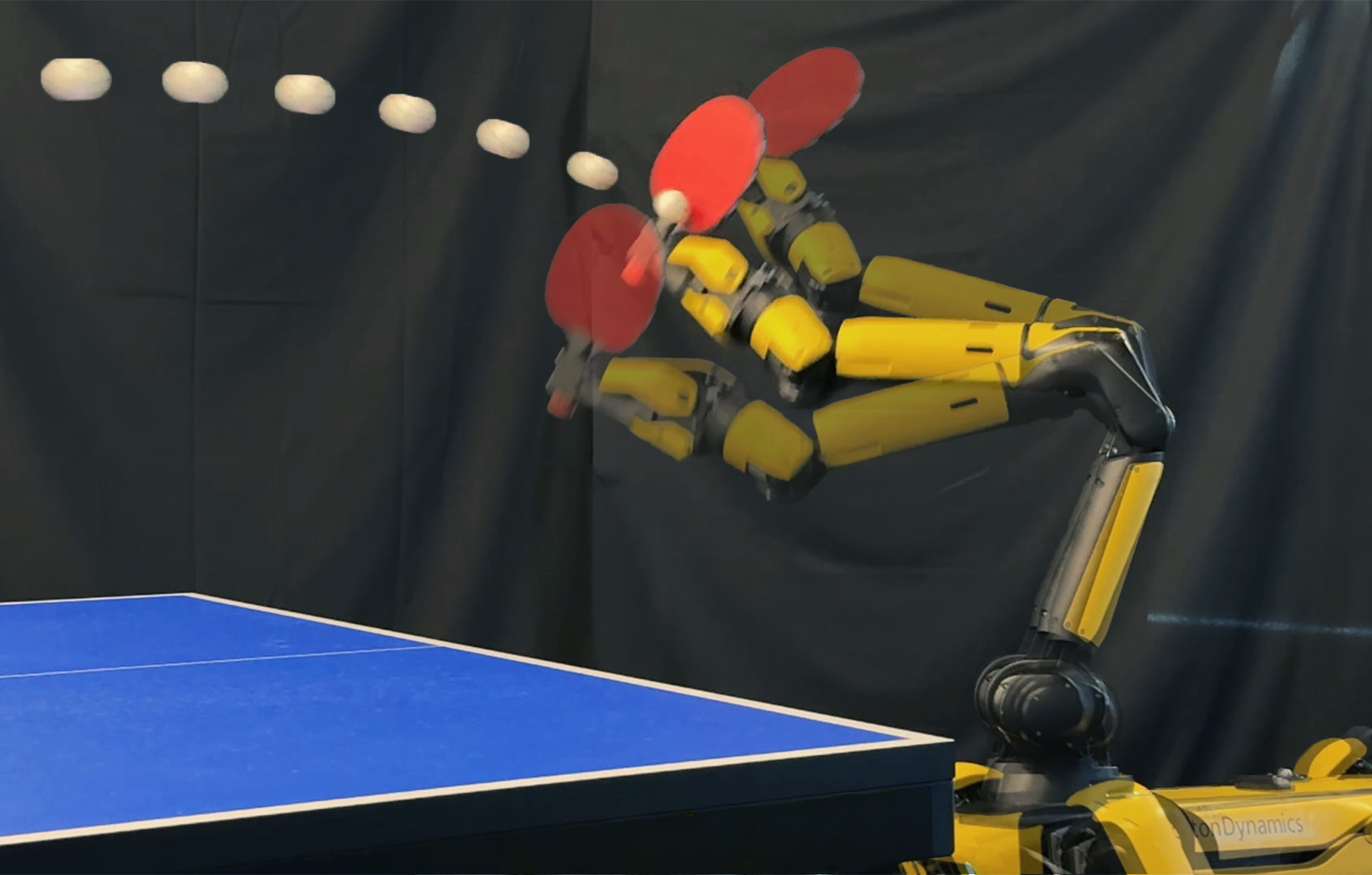}
         \label{fig:back_spin_hardware}
     \end{subfigure}
     \hfill
     \begin{subfigure}[b]{0.45\textwidth}
         \centering
         \includegraphics[width=\textwidth]{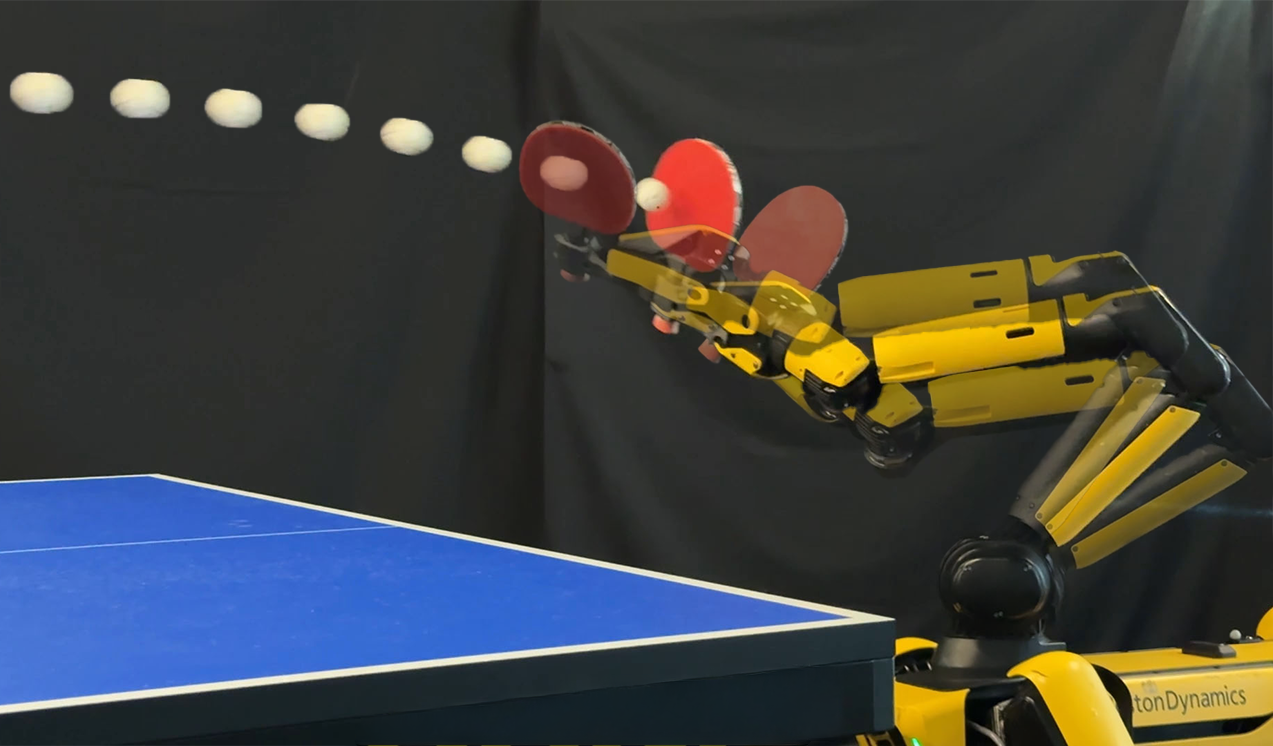}
         \label{fig:top_spin_hardware}
     \end{subfigure}
         \caption{Back spin (top) and top spin (bottom) hardware swing examples with the exiting ball trajectory shown.}
        \label{fig:hardware_swings}
\end{figure}

Over the four targets, the system is capable of getting within 20 rad/s of the desired $\boldsymbol{\omega}^+$ showing its ability to generate diverse spin. Examples of back spin and top spin shots are shown in Figure~\ref{fig:hardware_swings}.

\subsection{Table Tennis Game Play}

The quantitative evaluations above show how each sub-component operates well and that the integrated system is capable of handling spin as well as aiming. Qualitatively, our system also supports playing table tennis with a person and is able to sustain a rally of over 10 hits per player all while handling spin from the opponent.


\section{Conclusions}

We introduce a system that allows a quadrupedal robot equipped with a robotic arm to play table tennis. Our approach coordinates high-speed camera-based ball detection and localization, ball trajectory and spin prediction, aiming optimization, and whole-body model predictive control. This integration allows for agile and accurate table tennis play on a Spot robot. We evaluate our system on hitting accuracy, and demonstrate its ability to handle and generate ball spin. Future work aims to address the following limitations:

\subsubsection{Perception}
The current system relies on off-board cameras for ball detection. While prior work has developed onboard perception for slower racket sports such as badminton~\cite{eth_badminton}, extending this to table tennis which involves substantial rapid body movements remains a significant research challenge. 

\subsubsection{Prediction}
Human players often rely on observing the movements of their opponents to estimate ball spin and plan strategy. Developing a prediction pipeline that can extract useful information from motions of an opponent will also be an important part of future table tennis robots.

\subsubsection{Control}
A key limitation of our current controller is that it excludes stepping due to the challenges of real-time, contact-implicit optimization. To overcome this limitation, a hybrid approach combining a reinforcement learning-based method~\cite{g1_pingpong, eth_badminton} with our proposed MPC could enable the robot to perform the agile footwork essential in human table tennis, while still leveraging the MPC for efficient online swing planning.

\subsubsection{Strategy}
Currently, our stroke strategy, e.g., where to aim and return spin, is based on simple heuristics. To achieve a level of competitive play comparable to that of other table tennis robots such as~\cite{google_pingpong} or ranked human players, substantial research is required to develop more sophisticated and adaptive strategies.

\section{Additional Materials}
A supplementary video including explanatory
animations and hardware tests can be found at the following
link: https://www.youtube.com/watch?v=3GrnkxOeC14

\bibliographystyle{IEEEtran}
\bibliography{icra2026}


\end{document}